\documentclass{article}

\usepackage[preprint]{neurips_2024}

\usepackage[utf8]{inputenc} % allow utf-8 input
\usepackage[T1]{fontenc}    % use 8-bit T1 fonts
\usepackage{hyperref}       % hyperlinks
\usepackage{url}            % simple URL typesetting
\usepackage{booktabs}       % professional-quality tables
\usepackage{amsfonts}       % blackboard math symbols
\usepackage{nicefrac}       % compact symbols for 1/2, etc.
\usepackage{microtype}      % microtypography
\usepackage{xcolor}         % colors

\usepackage{amsthm,amssymb}
\usepackage{mathtools}

\usepackage[capitalize,noabbrev]{cleveref}
\theoremstyle{plain}

\theoremstyle{definition}

\title{A Theory of Machine Learning}

\author{%
  Jinsook Kim\thanks{Use footnote for providing further information
    about author (webpage, alternative address)---\emph{not} for acknowledging
    funding agencies.} \\
  Underwood International College\\
  Yonsei University\\
  Seoul, Korea 03722 \\
  \texttt{jki76364@gmail.com} \\
  \And
  Jinho Kang \\
  Emeritus, Seoul National University \\
  Seoul, Korea 08826 \\
  \texttt{jhkang@snu.ac.kr} \\
}
\begin{document}

\maketitle

\begin{abstract}
  We critically review three major theories of machine learning and provide a new theory according to which machines learn a function when the machines successfully compute it. We show that this theory challenges common assumptions in the statistical and the computational learning theories, for it implies that learning true probabilities is equivalent neither to obtaining a correct calculation of the true probabilities nor to obtaining an almost-sure convergence to them. We also briefly discuss some case studies from natural language processing and macroeconomics from the perspective of the new theory.
\end{abstract}

\section{Introduction}
 In this paper, we examine three major theories of machine learning. We will call them \emph{the possible worlds theory}, \emph{the recognition theory}, and \emph{the operation theory}. Both the possible worlds theory and the recognition theory are based on what we will call the \emph{epistemic} approach to machine learning, whereas the operation theory is based on what we will call the \emph{behavioral} approach. We will prove that all three theories have important problems. We will then provide a new theory of machine learning according to which machines learn a function when machines \emph{successfully} compute it. We will show that this theory challenges common assumptions in the statistical and the computational learning theories, for it implies that learning true probabilities is equivalent neither to obtaining a correct calculation of true probabilities nor to obtaining an almost-sure convergence to them. Lastly, we will discuss when machines can or cannot learn a probability function in the perspective of our new theory by considering two case studies, the first one from natural language processing and the second one from macroeconomics. 

\section{Two Theories of Machine Learning in the Epistemic Approach}

The epistemic approach to machine learning emphasizes that learning is the phenomenon of \emph{knowledge} acquisition. In order for machines to learn a function,  they must acquire knowledge of it. Given that what machines do is essentially computational, we can say that machines \emph{learn} when they acquire \emph{knowledge} through a \emph{computational} method. There are two influential theories of machine learning in the epistemic approach, the possible worlds theory and the recognition theory. Let us consider them in turn. 

\textbf{2.1 The Possible Worlds Theory}

First, the possible worlds theory: two things are defined here, (1) \emph{knowledge} and (2) the \emph{process} through which the learning algorithm returns the final knowledge from the completed task. For (1), knowledge is defined as \textit{truth in all epistemically possible worlds}. Based on this definition of knowledge, process is defined as follows: extending some notions from \cite{HalpernF:97,HalpernF:03}, let us define a computer system or process by the set of possible runs. Here, a \emph{run} is a description of the behavior of the system over time. Formally, a run is a function from time to state while each state encapsulates all the knowledge available to the system at that time. Now, this system may consist of either a single process or multiple processes. In the latter case, states can be further subdivided into local states and global states, and the system becomes a distributed system. It is worth noting here that time is discrete and does not need to be ``real time.'' To distinguish time from``real time'', we will call it a \textit{step}. Now, except for the initial knowledge base given in the initial state $s_{1}$, all the pieces of knowledge afterwards are \textit{internally} provided at each step while the algorithm is being executed. Thus, once any piece of knowledge is added internally to the state, an additional step is counted. Hence we can derive the following definition.

\textbf{Definition 1} Machines \textbf{learn} when their \emph{processes} are in the overall state $(s_{1},\ldots ,s_{n})$, where $%
s_{n}$ is the state at terminating step \textit{n} and each state $s_{i}$ encapsulates all pieces of \emph{knowledge}, i.e. every truth in all epistemically possible worlds, that are available at each step $i$ once their algorithms are executed. 

The possible worlds theory provides an excellent formalism for reasoning about knowledge. However, it does not explicate what learning is because it ultimately fails to explicate what knowledge is. That is, considering \cite{DeRose:91} and \cite{Stalnaker:06}, we derive the following theorem:

\textbf{Theorem 1} The definition of knowledge in the possible worlds theory does not provide any substantial analysis of how to determine the truth in all \textit{epistemically possible worlds} without relying on the very notion of knowledge.

If Theorem 1 holds, then the possible worlds theory cannot explicate what knowledge is in terms of truth in all epistemically possible worlds. Hence it cannot explicate what learning is either. Therefore, the possible worlds theory inevitably leads to the situation in which any knowledge-based programming must eventually rely on ``knowledge'' of the programming designers who design the program and set its state space. 

\textbf{2.2 The Recognition Theory}

We can point out a similar problem in the second theory of machine leaning in the epistemic approach, namely the recognition theory, which was proposed in \cite{Valiant:84}. According to \cite{Valiant:84}, machine learning can be defined as follows.

\textbf{Definition 2} Machines \textbf{learn} a concept if, given data, there exists a deduction procedure by which a correct \textit{recognition algorithm} is derived for that concept.

Here, \emph{algorithims} mean step-by-step instructions used in computing to correctly recognize the concept. So this theory is computational, which we will explain below. Just like knowledge, \emph{recognition} here is another epistemic notion that needs to be explicated in order to explicate what learning is. 

Rather than providing an explicit algorithm that would universally return correct recognition in every possible instance, the recognition theory in \cite{Valiant:84,Valiant:08} aims to control the error that machines fail to recognize a concept for given data. This is the point where the recognition theory is distinctive from the possible worlds theory: unlike the possible worlds theory, it does not aim to encode universal rules that can be applied to every possible case without fail. When errors are controlled arbitrarily small by increasing sample size at reasonable cost, machines are guaranteed to be robust with correct recognition. Note that the recognition algorithm is the Boolean circuit expressed by any concept function which is approximate to the true function, so that when the true concept function can be arbitrarily approximated by controlling error, the recognition algorithm can be deduced. What is ``recognition,'' then? In the case of concept recognition, it is what machines obtain by approximating the true target concept function. What is the ``true concept function,'' then? Machines do not need any explicit definition of the true concept function to recognize it because it would be enough if machines could approximate the function arbitrarily well. Technically, machines are guaranteed to control errors when errors come from the \textit{i.i.d.} distribution. In the end, then, \textbf{recognition} is whatever springs up to machines under the \textit{i.i.d.} assumption after executing some Boolean circuit by processing sufficient data. 

The recognition theory is a great breakthrough given the problem of induction that machines otherwise must face to learn universal rules from finite samples. However, the \textit{i.i.d.} assumption here cannot be innocent. It is one thing that machines need some \textit{i.i.d.} assumption to learn, while it is another that machines indeed satisfy such an assumption so that they learn. We thus derive the following theorem.

\textbf{Theorem 2} To satisfy the \textit{i.i.d.} assumption, the deducing recognition algorithm must rely on \emph{learning} a certain identical distribution.

Theorem 2, however, implies that the definition of learning in the recognition theory is ultimately based on the very notion of learning. Hence the definition is circular and cannot explicate what learning is.

\section{A New Theory of Machine Learning and Its Implications}

We have shown how formal treatments of epistemic notions such as ``knowledge'' in the possible worlds theory and ``recognition'' in the recognition theory fail to explicate what learning is. Thus, our strategy to define ``learning'' will be rather indirect, without committing ourselves to what knowledge or recognition is. Note that the phenomenon of learning must be at least computational in its essence when achieved by machines. Therefore, we adopt the notion of computation to define what learning is in Definition 3. Inspired by the ideas of \cite{Turing:36} and \cite{Church:36}, we require that machines be able to \emph{effectively calculate} or \emph{compute} a target function when machines can learn the function. Thus, once machines \emph{learn} a function, there must \emph{exist} some definite and explicit \emph{instructions} that machines must follow to return the function, which is to capture the role of ``machine learning'' as a computational notion.

\textbf{Definition 3} Machines \textbf{learn} any target function when they \textit{succeed} in effectively calculating/computing the function if any, after processing possibly infinite amounts of data.

It should be noted that we added the notion of success in Definition 3, which is to capture the role of ``learning'' as an \emph{epistemic} notion. The epistemic notion of machine learning requires two important components: if machines learn, (\textit{i}) they are $\textit{indeed correct}$ most of the time, and (\textit{ii}) they are $\textit{self-assured to be correct}$ most of the time. In other words, machines successfully compute a target function if and only if they compute it in a \textit{reliable} and \textit{doxastic} manner. As a result, we derive the following success criterion:

\textbf{The Success Criterion for Machines to Learn a target function} 

(1) If machines achieve computational success by learning, what they compute in the end must be at least true to our world most of the time, which should be assured to the machines themselves. 

If what they compute turns out to be wrong or the machines admit errors repeatedly too often out of infinite opportunities to learn, then their computation cannot be considered successful. In case of learning the true probability, we prove what exactly is meant by ``most of the time'' in the proof of Theorem 3. Furthermore, \cite{Kim:24} proves in Corollary 4.37 that Success Criterion (1) is a sufficient condition for learning in the case of computing the true probabilities. Thus, although the notions of ``computation'' and ``success'' are informal in Definition 3, we can provide under what formal conditions machines can achieve such computational success once machines are under the specific task of learning a particular function. 

Now that we define what ``machine learning'' is, let us discuss when machines can learn the true probabilities. Before doing so, let us first derive a necessary condition for machines to learn a true probability function.

In \cite{Hintikka:62}, the knowledge of a person $i$ refers to the knowledge of that person $i$ on any proposition $A$. Likewise, machine's learning of the true objective probability $P$ here refers to the knowledge acquired by any machine on the probabilistic proposition $A_{p}$. If machines learn the true probability as $\alpha$, then the probabilistic proposition $A_{p}$ amounts to that the true objective probability $P,$ if any, is what the very machines calculate as $\alpha$. Here, we convert the non-propositional learning into propositional learning. But it deserves to emphasize that machines do not lose any information in the representational content of learning while converting the non-propositional structure into propositional one. (e.g. \cite{Peacocke:06}) 

Now, just as a person $i$'s knowledge on proposition $A$ must satisfy the necessary condition that the person $i$'s belief in $A$ be true, machine learning of the true probability $P$ must also satisfy the necessary condition that the representation of $A_{p}$ of the machines be true. Note here that such a representation of $A_{p}$ is true when what has been computed by the machines is indeed equal to the true probability $P$. Now, this computed probability function by machines is nothing more than the subjective probability of the machines. Therefore, a necessary condition for machine learning of true probability $P$ requires that the subjective probability of machines is indeed in congruence with the true objective probability $P$. In short, if machines learn the true objective probability $P$, then the subjective probability $\Pi$ of the machines is actually equal to the true probability $P$.

\textbf{A Necessary Condition for Machines to Learn the True Probability}

(2) If machines \textit{learn} the true objective probability $P(A_{t+1}|$\ss $_{t})$, then $\Pi(A_{t+1}|$\ss $_{t})=P(A_{t+1}|$\ss $_{t})$ 

\ \ \ \ where
$\Pi(A_{t+1}|$\ss $_{t})$ denotes the subjective probability of the machines at time $t$. 

We assume, without loss of generality, that the event $A_{t+1}$ is an elementary event, for simplicity. Thus, the event $A_{t+1}$ is a singleton, i.e. $\{\omega_{t+1}\}$ where $\omega_{t+1}$ is a state in the sample space of the true probability. 

It is worth emphasizing that learning implies obtaining a true fact by satisfying the truth condition that $\Pi(A_{t+1}|$\ss $_{t})=P(A_{t+1}|$\ss $_{t})$, but that the converse does \textit{not} hold, which will be demonstrated shortly by constructing a counterexample. This is why the above is only a necessary but not sufficient condition for learning the true probability. In particular, we will prove that learning is not equivalent to any of the following three mathematical conditions, although such equivalence is commonly assumed in large literature. 

\textbf{3.1. Learning Is Not a Correct Calculation of True Probabilities} 

First, machine learning of true probabilities is not equivalent to a correct calculation of them by machines. That is, it is \emph{not} necessarily the case that machines learn even when they return a correct calculation that $\Pi(A_{t+1}|$\ss $_{t})=P(A_{t+1}|$\ss $_{t})$. If machines return correct calculations of true probabilities, then machines must have obtained true facts about some objective uncertainties of our world. But not every obtainment of a true fact is an acquisition of \emph{knowledge}, for machines may have \textit{happened} to compute the true probability only by \textit{luck}. Ever since \cite{PlatoM:14}, it has been accepted that merely an accidental coincidence with the truth by luck is not enough to count as knowledge. Since knowledge is not acquired through obtaining truth by luck, machines do not necessarily learn even if they obtain a true fact. 

However, there is large literature whose common assumption is that (acquired) knowledge is equivalent to a correct calculation of true probability. (e.g. \cite{BlumeE:06}, \cite{Sandroni:00}, \cite{SargentC:08}) In particular, this assumption justifies some game settings where certain winning strategies in games are equivalent to learning strategies in the theory of learning through games. (e.g. \cite{FosterV:93}, \cite{Nisan:07}) But such an assumption is challenged by the following Theorem 4.

\textbf{Theorem 3} If machines learn the true probability $P(A_{t+1}|$\ss $_{t})$ as $\alpha$, then $P $ $ (p_{k}\rightarrow$ $\alpha) =1$ with $k\to\infty$ where $k$ is the number of days in the test set,

\ \ \ \ \ \ \ \ \ \ \ \ $p_k=(
{\textstyle\sum\limits_{t=0}^{k-1}}
\xi_{t+1})^{-1}\cdot(
{\textstyle\sum\limits_{t=0}^{k-1}}
\xi_{t+1}\cdot 1_{\{A_{t+1}\}})$ with $1_{\{A_{t+1}\}}$ being the indicator of $A_{t+1}$ and

\ \ \ \ \ \ \ \ \ \ \ \ $\xi_{t+1}:=\begin{cases} 1 & $if $\Pi(A_{t+1}|$\ss $_{t}) = \alpha = P(A_{t+1}|$\ss $_{t}) \\ 0 & $if $\Pi(A_{t+1}|$\ss $_{t}) = \alpha \neq P(A_{t+1}|$\ss $_{t}) \end{cases}$ 

In general, a test set denotes a set whose elements are collected by certain selection criterion to test calibration property in \cite{Dawid:82}. Here, our selection criterion to construct a test set is a \textit{correct} probability as $\alpha$. 

\textbf{Theorem 4} If $P(A_{t+1}|${\ss} $_{t}) \neq \alpha $ at least for infinitely many $t$'s with $t \ge n$ for some $n < \infty$ along the stochastic path, then $P(A_{t+1}|${\ss }$_{t})=\alpha $ at some $t^{\ast} < n$ is not equivalent to \textit{learning} the true probability at $t^{\ast}$.

For example, we can think of the situation in Theorem 4 as an analogue to the following: even a broken clock is correct twice a day. But machines cannot be said to learn what time it is from the broken clock even when they happen to return the correct time, say 2pm, by processing the data from this clock luckily at the right timing, namely at 2pm. Therefore, ``learning'' should be distinguished from ``obtaining the true fact''. 

\textbf{3.2. Learning Is Not an Almost-Surely Correct Calculation of True probabilities} 

Second, machine learning of true probabilities is not equivalent to an almost-surely correct calculation of them by machines. That is, it is \emph{not} necessarily the case that machines learn even when they return an almost-surely correct calculation.

\textbf{Theorem 5} If $P($ $P(A_{t+1}|${\ss} $_{t}) \neq \alpha $ at least for infinitely many $t$'s with $t \ge n$ $) >0$ for some $n < \infty$ along the stochastic path of the test set, then $P($ $P(A_{t+1}|${\ss }$_{t})=\alpha $ $) = 1$ at some $t^{\ast} < n$ is not equivalent to \textit{learning} the true probability at $t^{\ast}$.

\textbf{Remark 1} We provide more detailed explanation on Theorem 5 with an example in the Appendix B.

However, there is large literature in which the condition of the true probability $P-$one is regarded as the definition or the necessary and sufficient condition for learning/knowledge. For example, \cite{AumannB:95} stipulates that the player $i$ is said to \textit{know} an event at state $s$ if he/she ascribes probability 1 to that event at $s$. It is not clear whether or not \cite{AumannB:95} meant by "probability" here the true objective probability. But since the condition of \textit{subjective} probability being 1 is usually interpreted as \textit{full confidence} but ``knowing'' is not necessarily the same as ``being fully confident/self-assured'', we assume that the probability in this context should be the true objective one, not just the subjective one. Let us then discuss how \cite{AumannB:95} can be understood in our context. 

Note that knowledge in the context of \cite{AumannB:95} is often knowledge about the player \textit{himself / herself}. For example, Lemma 2.6 in \cite{AumannB:95} says that player $i$ knows \textit{that he/she attributes probability $\pi$ to an event $E$} if and only if he/she indeed attributes probability $\pi$ to $E$ with the true probability $P-$ one. \cite{AumannB:95} establishes this Lemma 2.6, showing that if the player $i$ indeed attributes this, then with the true probability $P-$ one, the player $i$ does so. Now, note that whether the player attributes a certain (subjective) probability $\pi$ to $E$ or not is an \textit{internal} event that occurs to the player himself / herself, but that whether such an internal event occurs or not is a matter of fact. In any case, the definition of knowledge in terms of obtaining a true fact in \cite{AumannB:95} conflicts with our general argument that not every obtainment of a true fact is knowledge. In other words, Lemma 2.6 conflicts with our general argument that satisfying the truth condition with true probability $P-$ one is just a necessary condition for learning/knowing, not a sufficient one. We thus conclude that the way \cite{AumannB:95} uses ``knowledge'' is erroneous \textit{in general}. 

However, we should point out that humans' \textit{self-knowledge} has some privileged status. (e.g. \cite{Gertler:21, Gertler:11}) That is, humans' true beliefs about themselves are \textit{guaranteed} to be true in many cases so that they do amount to knowledge. So if humans are guaranteed to be true about themselves with true probability $P-$ one, they are not only fully confident/self-assured about themselves but also know about themselves. Hence human beings come to know that an internal event occurs to themselves whenever what is internally occurring to them holds with true probability $P-$ one. But more discussions of the privileged status of human self-knowledge are beyond the scope of this paper, so we will not investigate this issue in \cite{AumannB:95} any further. 

\textbf{3.3. Learning Is Not an Almost-Sure Convergence to True Probabilities} 

Third, machine learning of true probabilities is not equivalent to obtaining almost-sure convergence to them by machines. 

\textbf{Theorem 6} Even if machines correctly calculate that $P($ $\lim\limits_{t\rightarrow \infty }P(A_{t+1}|$\ss $_{t})=\alpha )=1$, this is not equivalent to that machines learn true probabilities $P(A_{t+1}|$\ss $_{t})$ as $\alpha$.

From the Success Criterion (1) and the necessary condition (2), we further derive that ``machine learning'' implies an almost sure convergence of the true probability point-wise, but not vice versa. As in the previous two sections, ``learning'' should be distinguished from obtaining `` the true fact'' that $P(A_{t+1}|$\ss$_{t})$ converges to $\alpha $ with true probability $P$-one. Note that \cite{Kim:24} explains by Lemma 4.31 why machines cannot learn even if they satisfy the condition of the almost-sure convergence: if machines are not self-assured of whether the true probabilities remain observable most of the time even when the true probabilities do remain so, machines cannot learn them. In other words, Corollary 4.37 in \cite{Kim:24} shows that even if machines satisfy the truth condition by calculating the true probability most of the time, they cannot be said to learn when they do not satisfy the second part of the Success Criterion that they must be self-assured of being correct most of the time when they indeed are correct that often. 

Now, note that while constructing a counter-example in the proof of Theorem 6, we show the following: consider an identically distributed true process $\mathbb{X}^{\ast}$, a sequence of random variables $\{X_t^{\ast}\}_{t=k}^{\infty}$ whose underlying common probability is $P$. Then, the data exclusively from this process $\mathbb{X}^{\ast}$ cannot be \textit{available} to machines point-wise, so that the machines cannot learn the common true probability distribution of the process $\mathbb{X}^{\ast}$. Machines cannot learn it because machines cannot determine the selection criterion $\xi_{t_k}$ of collecting data exclusively from $X_t^{\ast}$'s unless machines have been self-assured of what the true underlying probabilities are. As Theorem 4.36 and Corollary 4.37 in \cite{Kim:24} show, machines can be self-assured of what the underlying true probabilities are only when the true probabilities are directly observable because machines then can collect data from a \textit{given} population where an identical distribution is already endowed.

However, it has been widely accepted (e.g. \cite{Vapnik:00}) that almost sure convergence condition (e.g. Glivenko-Cantelli theorem) constitutes the definition of ``learning.'' But at least with regard to a true probability function, it is one thing that the functions of a sequence of random variables \textit{mathematically converge} to a certain limit almost surely, while it is another thing that machines can \textit{learn} the true function as such limit by processing the \textit{relevant} data. 

\section{The Operation Theory of Machine Learning in the Behavioral Approach}

Let us now consider the operation theory of machine learning, which is based on the behavioral approach, while discussing some fundamental limits on deriving practical implications from the theoretical results on learning true probabilities. Note that one way to prove a practical significance of any theoretical result on learning is to show that such a theoretical result implies some observational significance which can be quantitatively measured by some performance of real learning machines. According to \cite{Mitchell:97}, this idea succinctly constitutes the very definition of machine learning.

\textbf{Definition 4} A computer program \textbf{learn} from experience $E$ with respect to some class of tasks $T$ and performance measure $P$, if its performance at tasks in $T$, as measured by performance $P$, \textit{improves} with experience $E$. (\cite{Bengio:16})

Unfortunately, however, this definition cannot hold for the case of learning \textit{the true probability}, as long as the tasks of machines are in the form of standard models of optimal decision under uncertainty. For there exists an \textit{observational equivalence} in machine behaviors under various probability measures.

\textbf{Theorem 7}
Suppose that machines accomplish their tasks in the form of solving the standard optimization problems under uncertainty as follows:

\ \ \ \ \ \ \ \ \ \ \ \ max $\int f(x)$ $d\nu$ $=$ $\int f(x)h$ $d\mu$ \ \ \ \ \ \ subject to $x\in$ $\Game.$ 

\ \ \ \ \ \ \ where $f$ is any gain function and $h$ is the \textit{Radon-Nikodym} derivative of $\nu$ with respect to $\mu$.

Then, there exists observational equivalence in the optimal behaviors of machines under various probability measures, $\mu$ and $\nu$. Furthermore, suppose that the true probability is unique, that is, it is $\nu$ but not $\mu$. Then machines cannot learn the true probability under Definition 4.

Therefore, once we try to prove the practical significance of a theoretical result, we return to where we prove the impossibility of learning the true probability.

 We have called the theory in \cite{Mitchell:97} ``the operation theory'' in the sense that learning is defined by the operation of the subject to whom the learning occurs. The general idea behind the operation theory is that we can define some non-observational concepts such as ``learning'', a psychological concept, in the context of science only when we have a method of measurement for those concepts. This measurement is usually obtained by the outcomes of some specified manipulation through which the target operation is obtained in each case. The operation theory entitles scientists the right to do some empirical tests for the hypotheses formulated by non-observational terms. However, if there are always observational equivalences among the outcomes of multiple operations no matter what kind of manipulations are adopted, we cannot properly measure the outcomes of the target operation and, accordingly, cannot define those non-observational terms through any operation. Hence we conclude that the operation theory in general fails to explicate what learning true probabilities is, just as the possible worlds theory and the recognition theory fail to do so.

\section{Some Practical Implications for Machine Learning Algorithms}

However, once we abandon the idea of directly estimating the true probabilities by measuring machine performances which are quantitatively distinguished by learning different probabilities but let machines aim to directly estimate the true probabilities, there can be some possibilities to discuss practical implications from learning the true probabilities. 

\textbf{Theorem 8}
Let the machines aim at directly estimating the true probability itself. Then, there exists a parametric deep learning model which does not suffer from the problem of observational equivalence in Theorem 7.

Theorem 4.36 in \cite{Kim:24} proves under Definition 3 that ideal machines can learn the true probabilities if and only if the probabilities are \textit{directly observable} by the machines. Based on this Theorem 4.36, we will discuss two case studies, the first being the one in which real machines can learn true probabilities while the second being the one in which no real machines can do so. From these case studies, we can discuss what kind of practical implications for real machine learning algorithm we can derive from the theoretical results on ideal machine learning algorithm in \cite{Kim:24}. 

\textbf{Case 1. A Learnable Case from Natural Language Processing}

Deep learning has produced some astonishing results in speech recognition. (\textit{e.g.} \cite{Hinton:12}) This remarkable phenomenon can be chiefly explained by the fact that the true probability of obtaining any given sequence of words is indeed learnable by machines. Thus, as a simple example, let us consider the case of machine learning algorithms for a language model such as the N-gram model in the natural language processing. (\textit{e.g.} \cite{Jurafsky:08}) In particular, let us focus on simple (unsmoothed) N-grams as a baseline to understand the direct observation on the true probabilities in principle. For simplicity, we assume that the language we consider is English.

According to \cite{Jurafsky:08}, N-grams are essential in any task in which machines must identify words in noisy, ambiguous \textit{inputs}. Here, what deserves to note is the expression ``input.'' For example, in speech recognition tasks, machines are given the inputs in the form of ambiguous speech to be recognized. Then, once those inputs were ever used for actual speech and provided as what is needed to be further recognized by machines, there must exist some sort of \textit{true population} which contains whatever is the unambiguous form of the given input. This true population is what we call an \textit{ideal corpus} whose definition is provided below.

\textbf{Definition 5} At any given time $t_{0}$, let us consider the population $W^{\ast }$, the set of all words and sequences of words that have been used by any subject, such as real persons, for certain periods up to that time $t_{0}$ to\textit{ effectively} communicate each other in our world. We define the set $W^{\ast }$\ of all such words and sequences of words by the \textit{ideal corpus}.

For instance, Brown and Switchboard are some well-known corpora in practice. However, the corpus in Definition 5 is ideal in the sense that it includes all the relevant words and sequences of words that have ever been circulated within any community in this world for certain periods up to any given time. It is beyond any simple online collection of words and sentences. Here, words may also represent various ranges of things such as numbers, punctuation marks, fragments, filled pauses, tabs or spaces, etc. 

Thus, for any \textit{given} sequence of words $S$, what is the probability of obtaining $S$? This is the question that we would like to address here. This question, however, is different from what we would call the Markov question by following \cite{Markov:13}: What is the probability of obtaining any \textit{arbitrary} sequence of words? In the Markov question, we must consider a potentially infinite sequence of strings, and thus there does not exist any maximum size such as the maximum of word lengths or the maximum of size of the sequence of words in $W.^{\ast }$ It should be noted that the distinction between these two questions is particularly important because it implies that the way machines learn how to recognize sentences in a natural language is quite different from the way humans learn it, given that the ideal corpus $W^{\ast}$ in the simple N-gram model contains only a finite number of sentences whereas humans are supposedly able to learn how to recognize potentially infinite number of sentences in the natural language.

\textbf{Remark 2}  We provide more detailed explanation on the implications of the distinction between these two questions in the Appendix B.

Now, for any arbitrary finite number N $\in\mathbb{N}$, the unsmoothed N-gram models forecast the probability of obtaining the next word following the previous N-1 words for any \textit{given} sentence. This probability is in the form of conditional probability, which is connected to the joint probability of a sequence of words under the chain rule. The conditional probability then is defined to be measured by counting the frequencies in the ideal corpus in the following way:

$P(w_{n}|w_{1}^{n-1})=\frac{C(w_{n},w_{n-1},\ldots ,w_{1})}{C(w_{n-1},\ldots
,w_{1})}$ where $w_{1}^{n-1}$ represents the sequence of words $%
\{w_{1},\ldots ,w_{n-1}\}$ and $C(w_{1}^{n})$ counts the number of
frequencies that $w_{1}^{n}$ occurs in the \textit{ideal corpus}.

Then, by chain rule, $P(S)=P(w_{1}^{n})=\prod%
\limits_{k=1}^{n}P(w_{k}|w_{1}^{k-1})$ with $P(w_{1}^{0})=1$ for any given $S$ whose length is $n$.

Now, following \cite{Kim:24}, let us define when a true probability is directly observable and then prove that $P(S)$ is directly observable and thus learnable by machines.     

\textbf{Definition 6} Let us consider a set $W$ that consists of the sequence of events $A_{t+1}$'s, $\{A_{t+1}\}_{t=0}^{k-1}$ with $k$ potentially infinite. The set $W$ is then defined to be a \textbf{population} with $k$ number of elements, when this set $W$ is assumed to have a certain attribute of interest, and so an indicator variable $1_{\{A_{t+1}\}}$ is assigned to each event $A_{t+1}$ where $1_{\{A_{t+1}\}}$ has a value 1 or 0 depending on whether the event $ A_{t+1}$ satisfies such an attribute or not, once the set $W$ is collected. The empirical distribution of the population $W$ with respect to the given attribute is $\frac{1}{k}\sum\limits_{t=0}^{k-1} 1_{\{A_{t+1}\}}$.

\textbf{Definition 7} Machines \textbf{directly observe} $P$$(  A_{t+1}|${\ss }$_{t})$ from the \textit{true population} $W$ at $t^{\ast}$ if the following two conditions are satisfied: (\textit{i}) a true population $W$ is in principle \textit{available} to the machines. (\textit{ii}) machines effectively calculate the \textit{empirical distribution} of the population $W$ with respect to the given attribute, which is the true probability distribution of the event $A_{t+1}.$ 

Now, in case where the sequence $\{A_{t+1}\}_{t=0}^{k-1}$ is a time-series, Definition 7 means that $\Pi$$(  A_{t^{\ast}+1}|${\ss }$_{t^{\ast}})= \frac{1}{k}\sum\limits_{t=0}^{k-1} 1_{\{A_{t+1}\}}=P$$(  A_{t^{\ast}+1}|${\ss }$_{t^{\ast}})$ with $k=t^{\ast}$. Thus, if $t^{\ast}$ goes to infinity, then the directly observable true probability becomes the limiting relative frequency, the representative objective true probability. 

\textbf{Remark 3} We provide more detailed explanation on Definition 7 with an example in the Appendix B.

\textbf{Theorem 9} At any given time $t_{0}$, there exists the ideal corpus $W^{\ast }$ with a finite number of words and finite number of sequence of words that have been used by humans up to $t_0$. Then, for any given sentence $S$, the true probability of obtaining $S$ is effectively calculable from this corpus $W^{\ast }$ by machines. Furthermore, this true probability $P(S)$ is learnable by machines. 

In practice, facing various technological limits, real machines accomplish this task of learning the true probability by approximating through various methods such as maximum likelihood or some smoothing methods. Perhaps, real machines may approximate it by some other clustering algorithms to generalize the word N-gram to the class N-gram model while avoiding its distance dependency. No matter what approximation methods machines use, however, it is pertinent to show here that such a practical approximation can work well in principle with this case of directly observable probability, while it cannot work well in the other cases of not directly observable probability. We leave this issue of learning and approximation as a future research topic.

\textbf{Case 2. An Unlearnable Case from Macroeconomics and Finance}

Unlike what we have seen in Case 1, probability is not directly observable and so unlearnable in many cases. Recall by Definition 7 that machines can directly observe the true probability at least when there exists a true population available to machines. However, there is no population available to machines for r-star in macroeconomics and finance because r-star is a model-based estimate, not actual data, as large body of economics literature pointed out. (e.g. \cite{Krugman:23}) Therefore, in this unlearnable case, machines usually try to learn the true probability \textit{indirectly}, say by inferring it from other relevant observable variables while further assuming some system of the observation and the state equations that include unobservable variables with possibly known state-space. (e.g. Kalman Filter in \cite{Laubach:03})  

Now, we claim that no machines can estimate r-star because machines cannot learn the true probability of r-star nor approximate its true probability by learning the error/distance function of r-star. First, let us discuss some basic notations to define what the distance function is: let $(\Xi,$ \textbf{\ss }$)$ be a measurable space, where \textbf{\ss \ }is\textbf{ }a $\sigma-$algebra generated by random vector $X$ in $\Xi$. Now, call $\Xi$ the \textit{sample }space of $X$ and then fix a space $\Psi$, called the\textit{ value} space of $Y$. If there indeed exists a true function $f^{\ast}$ whose learning is a given task to machines, then a learning rule is any sequence of measurable functions $f_{n,m}:\Xi^{m}%
\times\Psi^{n}\times\Xi\rightarrow\Psi$ for $n,m\in%
%TCIMACRO{\U{2115} }%
%BeginExpansion
\mathbb{N}
%EndExpansion
\cup\{0\}$ where the domain of $f_{n,m}$ is the set of the following three things, $(i)$ the initial training samples of labeled data $(X_{1}%
,Y_{1}),\ldots,(X_{n},Y_{n})$ and $(ii)$ the unlabeled data $X_{n+1}%
,\ldots,X_{m},$ and $(iii)$ the random variable $X_{m+1}$ for prediction, given that $m>n$. Then, the output of $f_{n,m}$ is the prediction $Y$ of $X_{m+1}$ based on the given data. Thus, for example, if $m=n$ with $n>0$, then this learning rule is tied to supervised learning, while it is tied to semi-supervised learning with $m>n$ and $n>0$ and tied to unsupervised learning with $m>n$ but $n=0$. 

Now, let us define a function $\ell:$ $\Psi^{2}\rightarrow\lbrack0,\infty)$
, called the \textit{distance function }such that $\ell$ $($ $f_{n,m}(\{X_{i}\}_{i=1}^{m},$ $f^{\ast
}(\{X_{i}\}_{i=1}^{n}),$ $\{X_{i}%
\}_{i=m+1}^{\infty}),$ $f^{\ast}(\{X_{i}%
\}_{i=1}^{\infty}))$. In statistical learning theory, this distance function is usually called a loss function with $\ell$ $(y_{1},y_{2})=\ell$ $(y_{2},y_{1})$, which is connected to the notion of risk when the data come from \textbf{ i.i.d.}. Here, we do not require the distance function to be symmetric, because some distance function which measures the ``distance'' between the \textit{ probability} functions is not a metric. For example, the function $\ell$ here can include the Kullback-Leibler divergence. In general, this function $\ell$ is supposed to measure the distance between any two functions from $i)$ the predicted function $f_{n,m}$ for the sequence of future random vector $\{X_{i}%
\}_{i=m+1}^{\infty}$ which is obtained from the learning rule with the initial training samples and from $ii)$ its true function $f^{\ast}$ for $\{X_{i}\}_{i=m+1}^{\infty}$. 

\textbf{Theorem 10} Suppose that machines cannot learn the true probability function $P$ as $f_{\infty}^{\ast}$ even after processing infinitely many training samples $\{X_{i}\}_{i=1}^{\infty}$. Then, the machines cannot evaluate any $f_{n,m}$, which is a learning rule for any $n,m<\infty$, as a good (or bad) approximation to the true probability $f_{\infty}^{\ast}$.

Now that machines cannot obtain any good or bad approximation to the true probability of r-star because machines cannot learn its true probability, we conclude that machines cannot estimate r-star. We leave as future research topic what would be the alternative method of forecasting the evolution of macro-economy as time passes, when it is not possible to learn or approximate the true probability of such an important variable in the model.  

\section{Conclusion}
\label{sec:conc}
We close our paper with a few more remarks on how various theories of machine learning fare in the learnable case of Section 5 and the significance of the epistemic approach to machine learning.

First, we pointed out in Section 2 that the possible worlds theory of machine learning must rely on the knowledge of the program designer while setting the state space, and that the recognition theory bears the burden of proof that the distributional assumption is indeed satisfied. It seems hard to incorporate knowledge of the program designer into the system of the possible worlds theory unless it is first defined what knowledge is. Also, it seems difficult to show that the identical distribution assumption is satisfied in the system of the recognition theory unless it is first defined what learning is. In the cases of the directly observable probability, however, the true probability can be \textit{a priori} defined by the empirical distribution of the \textit{true population.} Since the target attribute of this population sets the state space of the true probability, it resolves the problem of incorporating  knowledge of the program designer in the possible worlds theory. Also, the true population shares the common identical distribution, so it satisfies the assumption of identical distribution in the recognition theory.    

Second, we emphasize that, in this paper, we have focused on the notion of ``machine learning'' which is not just computational but also epistemic, a counterpart to ``human learning.'' We have focused on this epistemic notion of machine learning because we particularly mean by ``machines'' some artifacts which perform \textit{human-level intelligent} behaviors. As long as we aim to contribute to the field of Artificial Intelligence through machine learning algorithms, machines must be such intelligent ``learners.''    

\bibliographystyle{chicago}

%\bibliography{references}

%%%%%%%%%%%%%%%%%%%%%%%%%%%%%%%%%%%%%%%%%%%%%%%%%%%%%%%%%%%%

\appendix

\section*{Appendix}

\section{Technical Proofs for Theorems}

\textbf{Proof of Theorem 1} Let an interpreted system $\mathcal{I}$ consists of a pair $(\mathcal{R}, \pi)$ where $\mathcal{R}$ is an system over a set $\mathcal{G}$ of states and $\pi$ is an interpretation over $\mathcal{G}$, which assigns truth values. Now that knowledge is defined as truth in all epistemically possible worlds, agent $i$ knows $\varphi$ if $\varphi$ is true at all points that agent $i$ considers epistemically possible. Thus, we define 

\begin{center}
(1) $(\mathcal{I}, r, m) \models \mathcal{K}_i \varphi$ if and only if $(\mathcal{I}, r', m') \models \varphi$ for all $(r', m')$ such that $(r, m) ~\sim_i (r', m')$ 

where $~\sim_i$ represents an epistemically possible relation to agent $i$.
\end{center}

However, due to \cite{DeRose:91},

\ \ \ \ (2) $(\mathcal{I}, r', m') \models \varphi$ if and only if $(\mathcal{I}, r, m) \nvDash \mathcal{K}_i \neg\varphi$

Therefore, we conclude that $(\mathcal{I}, r, m) \models \mathcal{K}_i \varphi$ in (1) cannot be determined without relying on whether $(\mathcal{I}, r, m) \nvDash \mathcal{K}_i \neg\varphi$ in (2), which shows that the notion of knowledge $\mathcal{K}_i$ in (1) by the agent $i$ relies on the very notion of knowledge $\mathcal{K}_i$ in (2) by the very agent $i$. 

In the possible worlds theory, \textit{knowledge} is standardly represented in terms of the possible worlds framework so that it is analyzed as \textit{truth in all epistemically possible worlds}. (e.g. \cite{Hintikka:62}, \cite{HalpernF:94}, \cite{Stalnaker:06}). Here \textit{epistemically} possible worlds must be distinguished from \textit{counterfactually} possible worlds. An epistemically possible world for an agent is the one that is compatible with the knowledge possessed by the agent, whereas a counterfactually possible world is any possible world that is compatible with all logical and mathematical truths. 

For example, let us suppose that Jane knows that it is Mount Everest that is the highest mountain in the world. Then there is no epistemically possible world for Jane, in which Mount Kilimanjaro is the highest mountain, because such a world is not compatible with what she knows. On the other hand, there are counterfactually possible worlds in which Kilimanjaro is the highest mountain, for Kilimanjaro’s being so is clearly compatible with all logical and mathematical truths. To put it another way, it is certainly the case that Kilimanjaro could have been the highest mountain so that we can easily imagine it without any contradiction. Sentences of the form ‘It is possible that P’, in which P is in the indicative mood, typically express epistemic possibilities, whereas sentences of the same form in which P is in the subjunctive mood typically express counterfactual possibilities. (e.g. \cite{DeRose:91}) Thus the sentence ‘It is possible that Kilimanjaro is the highest mountain’ typically expresses an epistemic possibility, whereas the sentence ‘It is possible that Kilimanjaro should have been the highest mountain” typically expresses a counterfactual possibility.

The above distinction between epistemically and counterfactually possible worlds shows that the analysis of knowledge in the possible worlds approach as truth in all epistemically possible worlds cannot be a reductive analysis, for we must use the very notion of knowledge in order to characterize the notion of epistemically possible worlds. $Q.E.D$
\bigskip

\textbf{Proof of Theorem 2} Let $X$ be a set which is called the \textit{instance space}. Then, a concept over $X$ is a subset $c \subseteq X$. Formally, a concept $c$ over $X$ is a boolean mapping $c: X \rightarrow \{0,1\}$, with $c(x)=1$ if $x\in c$ and $c(x)=0$ otherwise. Now, let $\mathcal{D}$ be any fixed probability distribution over $X$. Then, provided that $h$ is any hypothesis concept over $X$, the error between $h$ and the target concept $c$ can be defined as:

\begin{center}
$error(h) = Pr_{x\in\mathcal{D}}[h(x)\neq c(x)]$
\end{center}

Then, a \textbf{recognition} algorithm $L$ is deduced for the concept $c$ when, with probability at least $1-\delta$, the algorithm $L$ returns a hypothesis concept $h$ that satisfies $error(h)<\epsilon$ for every sufficiently small $\epsilon$ and $\delta$ and every probability distribution $\mathcal{D}$. Thus, no matter what distribution $x$ is drawn from, $x$ must be sampled from an \textit{identical} distribution. However, as we show by the counterexample in the proof of Theorem 6, collecting data from a fixed identical distribution can amount to ``learning'' the distribution itself. Therefore, in this case, deducing recognition algorithm must rely on the very notion of ``learning.'' $Q.E.D.$
\bigskip

\textbf{Proof of Theorem 3} For infinitely many $t$'s when
$P(A_{t+1}|$\ss $_{t})$ stays the same as $\alpha$ by Lemma 4.5 in \cite{Kim:24}, suppose that machines learn this $P(A_{t+1}|$\ss $_{t})$ as $\alpha$ at some time $t_0$. Then, by the Success Criterion (1), $\Pi(A_{t_k+1}|$\ss $_{t_k}) = \alpha = P(A_{t_k+1}|$\ss $_{t_k})$ at least infinitely often out of infinite opportunities at $t$'s to learn. (We prove in the below what we mean exactly by ``most of the time'' in the Success Criterion (1). Here we tentatively mean ``at least $i.o.$'' by it because machines are otherwise wrong too often to learn given the Success Criterion (1).) 

Thus we can construct a test set which consists of the subsequence of
$\Pi(A_{t_{k}+1}|$\ss $_{t_{k}})$ which is equal to $P(A_{t_{k}+1}%
|\text{\ss }_{t_{k}})$ for those infinitely many $t_{k}$'s$.$ Let $\xi
_{t_{k}+1}=1$ if and only if $\Pi(A_{t_{k}+1}|$\ss $_{t_{k}})=P(A_{t_{k}%
+1}|\text{\ss }_{t_{k}})=\alpha.$ Note that $\xi_{t_{k}+1}$ is \ss $_{t_{k}}%
-$measurable, because machine forecasting $\alpha$ occurs at time $t_{k}$. Then, by Theorem 4.1 in \cite{Kim:24} while replacing $\Pi$ by $P$, with true probability $P-$one, $p_{k}- $ $\alpha = (%
%TCIMACRO{\tsum \limits_{j=0}^{k-1}}%
%BeginExpansion
{\textstyle\sum\limits_{j=0}^{k-1}}
%EndExpansion
\xi_{t_{j}+1})^{-1}\cdot%
%TCIMACRO{\tsum \limits_{j=0}^{k-1}}%
%BeginExpansion
{\textstyle\sum\limits_{j=0}^{k-1}}
%EndExpansion
\xi_{t_{j}+1}(Y_{t_{j}+1}-\alpha)\rightarrow0$, as $k\rightarrow\infty$ where
$P$ is defined over \ss $_{\infty}=%
%TCIMACRO{\tbigvee \limits_{t_{k}=0}^{\infty}}%
%BeginExpansion
{\textstyle\bigvee\limits_{k=0}^{\infty}}
%EndExpansion
\text{\ss }_{t_{k}}$ and \ss $_{t_{k}}$ is denoted by the totality of
\textit{true} \textit{facts} up to day $t_{k}.$ Thus if machines learn the true probability $P(A_{t+1}|$\ss $_{t})$, then machines satisfy the calibration property by \cite{Dawid:82} along the stochastic path of the test set where machines are correct at least $i.o.$ out of infinite opportunities to learn. 

Now, by Lemma 4.10 and Theorem 4.17 in \cite{Kim:24}, machines cannot satisfy the calibration property when the test set is constructed by the selection criterion of an assessed probability $\alpha$ if $P(P(A_{t+1}|$\ss$_{t})\neq\alpha$ at least $i.o.) >0$. Therefore, in order to learn, the machines must return the correct calculations except a finite number of times out of infinite opportunities to learn. Thus, ``most of the time'' in the Success Criterion (1) should be ``all but finitely often out of infinite opportunities to learn,'' which means that machines must be correct not just infinitely often while being wrong that often. $Q.E.D.$
\bigskip

\textbf{Proof of Theorem 4} For any machine forecast $\alpha\in \mathbb{R} [0,1]$, suppose that $P(A_{t+1}|${\ss} $_{t}) \neq \alpha $ at least for infinitely many $t$'s with $t \ge n$ for some $n < \infty$ along the stochastic path but that $P(A_{t+1}|${\ss }$_{t})=\alpha $ at some $t^{\ast} < n$. Then, for $\alpha=1$, $P(P(A_{t+1}|$\ss$_{t})\neq\alpha$ at least $i.o. )>0$ for some event {$A_{t+1}$} with $t \ge n$ for some $n < \infty$ by Theorem 4.18 in \cite{Kim:24}. But $\{\omega\in\text{\ss }_{\infty}=%
%TCIMACRO{\tbigvee \limits_{j=0}^{\infty}}%
%BeginExpansion
{\textstyle\bigvee\limits_{t=0}^{\infty}}
%EndExpansion
\text{\ss }_{t}:$ $ 1_{\{\omega\}}=1$ when $P(A_{t+1}|\text{\ss }_{t})\neq 1$ for all $t \ge n$ for some $n < \infty\}\subset\{\omega\in\text{\ss }_{\infty}=%
%TCIMACRO{\tbigvee \limits_{j=0}^{\infty}}%
%BeginExpansion
{\textstyle\bigvee\limits_{t=0}^{\infty}}
%EndExpansion
\text{\ss }_{t}:$ $ 1_{\{\omega\}}=1$ when $P(A_{t+1}|\text{\ss }_{t})\neq \alpha$ for all $\alpha\in\mathbb{R}[0,1]$ and for all $t \ge n$ for some $n < \infty\}$. Therefore, for any $\alpha\in\mathbb{R}[0,1]$, $P(P(A_{t+1}|$\ss$_{t})\neq\alpha$ at least $i.o. )>0$ for some event {$A_{t+1}$} with $t \ge n$ for some $n < \infty$. Then, by (Case 3) of Theorem 4.16 in \cite{Kim:24} and Theorem 3, the machines cannot learn the true probability $P(A_{t+1}|$\ss$_{t}$ as $\alpha$. $Q.E.D.$
\bigskip

\textbf{Proof of Theorem 5} Suppose that $P($ $P(A_{t+1}|${\ss} $_{t}) \neq \alpha $ at least for infinitely many $t$'s with $t \ge n$ $) >0$ for some $n < \infty$ along the stochastic path of the test set. Then, by the Case 3 of Theorem 4.16 in \cite{Kim:24}, $P(p_{k}\rightarrow$ $\alpha)\neq1$ where $p_{k}$ denotes the limiting relative frequency along the path of the test set. Thus, by Theorem 3, machines cannot learn the true probability $P(A_{t+1}|${\ss} $_{t})$ as $\alpha$.

Now, as $P($ $P(A_{t+1}|${\ss} $_{t}) \neq \alpha $ at least for infinitely many $t$'s with $t \ge n$ $) >0$ for some $n < \infty$ by assumption, consider a special case where $P($ $P(A_{t+1}|${\ss} $_{t}) \neq \alpha $ at least for infinitely many $t$'s with $t \ge n$ $) = 1$. Then, even if $P($ $P(A_{t+1}|${\ss }$_{t})=\alpha $ $) = 1$ for some $t^{\ast} < n$, machines cannot learn the true probability $P(A_{t+1}|${\ss} $_{t})$ as $\alpha$. $Q.E.D.$ 

\textbf{Proof of Theorem 6} We prove this theorem by constructing a following counterexample where the almost-sure convergence to true probability function is satisfied but learning is not possible.

\textbf{Counterexample} Let us consider a simple process $\mathbb{X}$ whose realized values consist of the set $D$ = $\{x_{1}, \ldots, x_{k}\}$ for each random variable $X_{t}$ with a representative $x$ $\in$$D$. Here, the set $D$ is countably many, with $k$ potentially infinite, because the number of functional values that machines can calculate is countable, and so without loss of generality we assume that the random variable $X_{t}$ is discrete. 

Now, let us assume that the process $\mathbb{X}$ has the following underlying true probabilities along the stochastic path: for any $n \in \mathbb{N}$, on the [$\frac{n}{2}$] number of the periods $t'$s where $\{X_{t}=x\}$ occurs at each $t$ and [$\frac{n}{2}$] is the nearest integer to $\frac{n}{2}$, $P(X_{t}=x|$\ss $%
_{t-1})= \alpha$, while $P(X_{t}=x|$\ss $%
_{t-1})= \beta$ on the rest of the periods for any $x \in \{x_{1}, \ldots, x_{k}\}$ with $\alpha \neq \beta$ along the stochastic path. Then, provided that $\lim\limits_{n\rightarrow\infty}\frac{1}{n}\sum\limits_{t=1}^{n} 1_{\{X_{t}=x\}}$ exists with true probability $P-$one, 

\bigskip
(1) \ \  $P$ $(\lim\limits_{n\rightarrow\infty}\frac{1}{n}\sum\limits_{t=1}^{n}P(X_{t}=x|${\ss}$_{t-1})= \frac{\alpha + \beta}{2}) = 1$ if and only if $P$ $(\lim\limits_{n\rightarrow\infty}\frac{1}{n}\sum\limits_{t=1}^{n} \mathbf{1}_{\{X_{t}=x\}} = \frac{\alpha + \beta}{2})=1$ 

\bigskip
by Lemma 4.10 and Lemma 4.11 in \cite{Kim:24} (a similar result by Theorem 2.3.9 in \cite{Durrett:19}.) 

However, note that given $\alpha\neq\beta$, $\lim\limits_{n\rightarrow\infty}\frac{1}{n}\sum\limits_{t=1}^{n} \mathbf{1}_{\{X_{t}=x\}}$ returns neither of the two true probabilities $\alpha$ nor $\beta$. Now, $\frac{1}{n}\sum\limits_{t=1}^{n}1_{\{X_{t}=x\}}$ converges almost surely and $\frac{1}{n}\sum\limits_{t=1}^{n}P(X_{t}=x|$\ss $%
_{t-1})$ does so as well by (1). Also, any arithmetic average of probabilities is also probability, because it satisfies three Kolmogorov axioms. Let us denote this new probability by $P_n$ which is $\frac{1}{n}\sum\limits_{t=1}^{n}P(X_{t}=x|$\ss $%
_{t-1})$. Thus, the condition of almost-sure convergence is satisfied by the sequence of probability functions $\{P_n\}_{n=1}^{\infty}$ but machines cannot ``learn'' any true probability from this convergence condition. The machines were able to effectively calculate the limit value of the convergent sequence of $P_n$'s, but could not succeed in obtaining any true value, $\alpha$ or $\beta$.

Now, to learn the true probability $\alpha$ or $\beta$ from data along the stochastic path, machines must classify the process $\mathbb{X}$ according to its underlying true probabilities so that $\lim\limits_{n\rightarrow\infty}(%
%TCIMACRO{\tsum \limits_{t=1}^{k}}%
%BeginExpansion
{\textstyle\sum\limits_{t=1}^{n}}
%EndExpansion
\xi_{t})^{-1}\cdot%
%TCIMACRO{\tsum \limits_{t=1}^{k}}%
%BeginExpansion
{\textstyle\sum\limits_{t=1}^{n}}
%EndExpansion
(\xi_{t}\cdot 1_{\{X_{t}=x\}}) = \alpha$ with $\xi_{t}=1$ when $t$ comes from the periods when $P(X_{t}=x|$\ss $%
_{t-1})= \alpha$ and $\xi_{t}=0$ otherwise, and so on. Then, we obtain the following from (1),

\bigskip
(2) \ \ \ $P(\lim\limits_{n\rightarrow\infty}(%
%TCIMACRO{\tsum \limits_{t=1}^{k}}%
%BeginExpansion
{\textstyle\sum\limits_{t=1}^{n}}
%EndExpansion
\xi_{t})^{-1}\cdot%
%TCIMACRO{\tsum \limits_{t=1}^{k}}%
%BeginExpansion
{\textstyle\sum\limits_{t=1}^{n}}
%EndExpansion
(\xi_{t}\cdot 1_{\{X_{t}=x\}}) = \alpha) = 1$ if and only if $P$ $(\lim\limits_{n\rightarrow\infty}\frac{1}{[\frac{n}{2}]}\sum\limits_{k=1}^{[\frac{n}{2}]}P(X_{t_k}=x|${\ss}$_{t-1})= \alpha) = 1$
\bigskip

Now, without loss of generality, let $\mathbb{X}^{\ast}$ consist of the subsequence $\{X_{t_k}=x \}_{k=1}^{\infty}$ out of $\mathbb{X}$ where $t_k$'s are chosen when $\xi_{t_k}=1$. Therefore, to learn the underlying true probability of $\mathbb{X}^{\ast}$ as $\alpha$, the machines must determine whether the selection criterion of $\xi_{t_k}=1$ is satisfied or not at each $t$ along the stochastic path. However, to determine whether $\xi_{t_k}=1$ or not, machines must have learned whether $P(X_{t}=x|$\ss $%
_{t-1})= \alpha$ or not at each $t$. Then, this ends up with the result that machines must have learned the true underlying probability, in order to learn the true underlying probability. Clearly, learning is impossible in such a viciously circular way, no matter what definition of ``learning'' is adopted, including Definition 3 in this paper. 

Therefore, we conclude that although the almost sure convergence condition is satisfied by this process $\mathbb{X}^{\ast}$, learning the true probability of this process $\mathbb{X}^{\ast}$ is not possible for machines. We prove by this counterexample that learning is not equivalent to the condition of almost-sure convergence. $Q.E.D$
\bigskip

\textbf{Proof of Theorem 7} Suppose that machines solve a simple optimization
problem in $\Re^{n}$ where the expected values of a given function $f:$
$\Re^{n}\rightarrow\Re$ are to be maximized over a given set $\Game\subset$
$\Re^{n}$. The function $E$ $[f(x)]$ is called the objective function, say the
expected gain function in the standard optimization model, and the set $\Game$
the constraint set. Now, let us suppose that $\mu$ and $\nu$ are two different
probability measures such that $\nu$ is absolutely continuous with respect to
$\mu$. Then, there exists the \textit{Radon-Nikodym} derivative $h$ of $\nu$
with respect to $\mu$ so that the optimization problems under these two
measures can be expressed as follows:

\bigskip

\ \ \ \ \ \ \ \ \ \ \ \ max $\int f(x)$ $d\nu$ $=$ $\int f(x)h$ $d\mu$ subject
to $x\in$ $\Game.$
\bigskip

Therefore, unless $h$ is an identity function, the machines solve two different problems so that they optimize two different gain functions, $f(x)$ and
$f(x)h$ under two different probability measures $\nu$ and $\mu$. Notwithstanding, since the first-order conditions of the optimization problem
are the same as $\frac{d}{dx}\int f(x)$ $d\nu$ $=$ $\frac{d}{dx}\int f(x)h$
$du=\eta\cdot\frac{d}{dx}g(x)$ where $\Game=\{x|$ $g(x)=0\}$ and $\eta$
denotes a Lagrangian multiplier, we will obtain the same maximizer $x^{\ast}$ as a solution, if any. Thus, we obtain the observational equivalence in the machine behaviors which manifest themselves as the same maximizer under different probability measures. 

Now, it is equally possible that observationally equivalent behaviors, i.e., what are manifested as the same maximizer $x^{\ast}$, are derived from learning the true measure $\nu$ or some other measure $\mu$. Therefore, under observational equivalence, we cannot exclusively measure machine performance by learning the true probability $\nu$ rather than some other probability $\mu$, because:

\ \ \ \ \ provided that $h$ is not an identity function and two measures, $\nu$ and $\mu$, are different but absolutely continuous each other, $\int f(x^{\ast})$ $d\nu$ $=$ $\int f(x^{\ast})h$ $d\mu$.

Then, since there is no way to measure any improvement or worsening in machine performance exclusively from learning the true probability $\nu$ rather than learning the probability $\mu$, we cannot show the practical significance of learning the true probability $\nu$ under this framework, which again implies that machines cannot learn the true probability by the very definition of \cite{Mitchell:97}, Definition 4. $Q.E.D.$
\bigskip

\textbf{Proof of Theorem 8} For random vectors $Y_t \in\mathbb{R}$ and $X_t \in\mathbb{R}^{I}$, let $y$ and $x$ denote generic realizations and $y_t$ and $x_t$ denote sample realizations at $t$. Now, for some $s<t,$ let $f(Y_t | \{ X_i\}_{i=s}^{t})$ be the conditional probability $P(Y_t = y| \{ X_i = x\}_{i=s}^{t})$ from the true data-generating distribution and let $f_m(Y_t | \{ X_i\}_{i=s}^{t})$ be our parametric model $P_{m}$ for the true probability $P$. Then, given that $Y$ and $X$ represent some data sequences of $\{Y_t = y_t\}_{t=1}^{n}$ and $\{X_t = x_t\}_{t=1}^{n}$, we can rewrite the optimization equation in Theorem 7 by the joint conditional probabilities as follows:

\bigskip

(1) \ \ \ \ \ \ \ \ \ \ \ \ \ \ \ \ \ \ min $\int \frac{f(Y|X)}{f_m(Y|X)}$ $= \frac{P(Y|X)}{P_m(Y|X)}$ $dP$ 
\bigskip

Here, note that this is the problem of minimizing the Kullback-Leibler divergence. Then, under the assumption that our probability model is the ``closest'' to the true data-generating function, machines can estimate the parameters of our model by solving this minimization problem. Now that the minimizing solution, if any, will be the same as in the following problem, we can rewrite (2) as follows;

\bigskip

(2) \ \ \ \ \ \ \ \ \ \ \ \ \ \ \ \ \ \ max $\int log$ $P_m(Y|X)$ $dP$ 
\bigskip

Then, for example, under the assumption that the logarithmic representation of our model likelihood function is concave with respect to the parameters, machines can estimate the parameters from the given data $X$ and $Y$ using the deep learning method. Therefore, given $I$ number of training samples $X_{ti}$'s at time $t$ and $J_{k}$ number of latent variables $Z_{tj}$'s in each $k$-th layer with $K$\ number of layers, machines now calculate the following equation:

\bigskip

(3) \ \ \ \ \ \ \ \ \ \ $P_m(Y_{t}$ $|$ $X_{t};\widehat{\Theta})= P_m(Y_{t} ; g(Z_{t1}^{K},\ldots,Z_{tJ_{K}}%
^{K};\widehat{\Theta}_{K}))$ 

\bigskip

where $Z_{tj_{K}}^{K}=h^{(K)}(Z_{t1}^{K-1},\ldots,Z_{tJ_{K-1}}^{K-1}%
;\widehat{\Theta}_{K-1})$ for $j_{k}=1,\ldots,J_{K}$ and $Z_{tj_{1}}^{1}=h^{(1)}%
(X_{t1},\ldots,X_{tI};\widehat{\Theta}_{1})$ for $j_{1}=1,\ldots,J_{1}$ and $\widehat{\Theta}_{i}$
denotes the set of estimated parameters at each layer $i\in\{1,...,K\}$.
\bigskip

Note that we can obtain the optimization problem (2) from the setting of Theorem 7 once we replace the objective function $f$ in Theorem 7 by the log of some \textit{given} parametric function $P_m(Y|X)$. Then, the problem of observational equivalence in disappears now, because $f(x)$=$f(x)h$ in Theorem 7 is uniquely fixed as the log of $P_m(Y|X)$ with respect to the true measure $P$ and therefore $h$ must be an identity function. $Q.E.D$
\bigskip

\textbf{Proof of Theorem 9} Let each word $w$ with a length $n_{w}$\ be a sequence of the letters of the alphabet $\{a_{1},\ldots,a_{n_{w}}\}$ and each sentence $S$ with a length $n_{s}$ be a sequence of words $\{w_{1},\ldots ,w_{n_{s}}\}$ that humans have circulated to effectively communicate up to a certain time $t_0$. Here, sentence is distinguished from arbitrary combination of ungrammatical sequence of words that cannot be used for human communication. We can extend the set of the letters of the alphabet to include all the punctuation marks, numerals, etc., if necessary. Note that for any $w\in W^{\ast }$, $n_{w}$ must be finite, because no word can be included in the ideal corpus if its sequence goes to infinity. This is so because any infinite sequence cannot actually be used to effectively communicate at any given time. For the same reason, for any $S\in W^{\ast }$, $n_{s}$ must be finite as well, except for the cases where some
words are contained in sentences repeatedly infinitely often. In such exceptional cases, using the expression ``$\ldots$'' which can be treated as the 27th alphabet, we let those sentences contain the word which consists of unitary sequence of this 27th alphabet. Then, for any $t_{0}$, there must exist some fixed number $%
n_{w}^{\ast }$ and $n_{s}^{\ast }<\infty $ such that $n_{w}^{\ast }$ denotes the the maximum size of the sequences of alphabets among all the words in the ideal corpus $W^{\ast }$ while $n_{s}^{\ast }$ denotes the maximum size of the sequences of words among all the sentences in $W^{\ast }$. 

Now, note that the size of the population $W^{\ast }$ must be less than $\lambda \times $ $($27 to the power of $n_{w}^{\ast }\times
n_{s}^{\ast })$ for some $\lambda \in 
%TCIMACRO{\U{2115} }%
%BeginExpansion
\mathbb{N}
%EndExpansion
$ as follows: here, $\lambda $ is fairly large but countable, because the number of frequencies for actual usages of any given natural language is at most countably many, given that there are at most countable number of people who have used those languages to effectively communicate throughout history up to a given time $t_{0}<\infty $ in our world. Now, let us denote the population with size $\lambda \times $ $($27 to the power of $n_{w}^{\ast }\times
n_{s}^{\ast })$ by $W_{\max }.$ Then, at any given time $t_{0}$, there always exists the true population $W^{\ast }\subseteq $ $W_{\max }$ which contains the accumulated history of the words or sentences that had been circulated up to $t_{0}$. The size of $W^{\ast }$ is smaller than that of $W_{\max }$, because not all possible sequences of words in $W_{\max
} $ are grammatical, so they may not be circulated for actual usage to effectively communicate. Now, if $%
W^{\ast }$ does not exist within $W_{\max }$, it implies that there must exist some sentence $S^{\ast }$ with the size greater than $n_{s}^{\ast }$
such that $S^{\ast }\in W^{\ast }$, which is contradictory, because $%
n_{s}^{\ast }$ is the maximum size of all the sentences in $W^{\ast }$. 

Now, let us denote a function $f_{c}$ from the population $W^{\ast }$ to the set of natural numbers by a counting function. This counting function $f_{c}$ counts the frequencies of each given sentence $S^{\ast }$ out of the entire true population $W^{\ast }$. Since the set of languages is lexicographically Turing-enumerable, the counting function $f_{c}$ is Turing machine calculable. Now that $W^{\ast }$ exists and that the counting function $f_{c}$ is Turing-machine calculable, machines can effectively calculate the relevant empirical distribution out of the population $W^{\ast }$, which is the true probability distribution because the counting function $f_{c}$ provides $P(w^{k}|w_{1}^{k-1})$ for each $k$ with $1\leq k\leq n_{s}$ and thus provides $P(S)$ by chain rule to machines for any given $n_{s}$. 

Now, the true population $W^{\ast }$ is available in principle to machines at any given time $t_{0}$ because the size of $W^{\ast }$ is finite while the machines are ideal ones in the sense that they can disregard any practical limits to process data. Also, the machines can effectively calculate the empirical distribution of this $W^{\ast }$, which returns the true probability $P(S)$ because $P(S)$ is \textit{defined} to be $P(w_{1}^{n})=\prod%
\limits_{k=1}^{n}\frac{C(w_{k},w_{k-1},\ldots ,w_{1})}{C(w_{k-1},\ldots,w_{1})}$ where $C(w_{0})$ denotes the number of all the words in the ideal corpus. Therefore, $P(S)$ is directly observable to machines by Definition 7. Finally, by Theorem 4.36 in \cite{Kim:24}, $P(S)$ is learnable by the machines. $Q.E.D.$ 
\bigskip

\textbf{Proof of Theorem 10} Let us suppose $f_{\infty}^{\ast}$ exists and further that $f_{\infty}^{\ast}=P$ where $f_{\infty}^{\ast}=f^{\ast}(\{X_{i}%
\}_{i=1}^{\infty})$ and $P$ is the true probability of r-star. Then, by Definition 7, the true probability of r-star is not directly observable because r-star is a model-based estimate, not actual data. Thus, machines cannot learn the true probability of r-star by Theorem 4.36 in \cite{Kim:24}. If $f_{\infty}^{\ast}$ does not exist, then machines trivially cannot learn the true probability of r-star becasue there exists nothing to learn. Now that machines cannot succeed in effectively calculating $f^{\ast}$ by Definition 3, $\ell$ is not well defined for machines because one of the arguments of the distance function $\ell$ is not calculable by machines. Therefore, the machines cannot evaluate whether any learning rule $f_{n,m}(\{X_{i}% 
\}_{i=1}^{m},$ $f^{\ast}(\{X_{i}\}_{i=1}^{n}),$ $X_{j})$ is within any bound from $f^{\ast}(X_{j})$ for any $j\geq m+1$, for any $n,m<\infty$. $Q.E.D.$

\section{Further Detailed Remarks}

\textbf{Remark 1} For example, if the following equation (2) holds, then $P($ $P(A_{t+1}|${\ss}$_{t}) = \alpha ) = 1$ is not equivalent to that machines learn true probability $P(A_{t+1}|${\ss}$_{t})$ as $\alpha$ at $t$.

(2) \ \ \ \ $P$( $P(A_{2}|${\ss}$_{1})=\alpha, \ldots ,$ $P(A_{n}|${\ss}$_{n-1})=\alpha$, $P(A_{n+1}|${\ss}$_{n}) \neq \alpha$, $P(A_{n+2}|${\ss}$_{n+1}) \neq \alpha,\ldots )=1.$ 

At first glance, this result may look puzzling, because machines are not said to learn even if they correctly calculate the true probability for the first finite $n-1$ number of times with true probability $P$- one. However, recall the discussion on the necessary condition (1) that obtaining the true probability \textit{by luck} cannot be considered as learning. In fact, according to equation (2), here it is not only the case that $P($$P(A_{2}|${\ss} $_{1})=\alpha, \ldots ,$ $P(A_{n}|${\ss} $_{n-1})=\alpha)=1$, but also the case that $P($ $P(A_{n+1}|${\ss} $_{n}) \neq \alpha,$ $P(A_{n+2}|${\ss} $_{n+1}) \neq \alpha,\ldots )=1$. Now, note that if the machines had not been calculating the true probability correctly only by luck, they could not have been correct only for the \textit{finitely} many times and wrong for the rest of the \textit{infinite} times, with true probability $P$- one. For this reason, learning is defined as a \textit{success} in effective calculation in Definition 3. Calculating the correct values only for some finite times by luck while wandering around cannot be considered as a genuine success. If such computations had indeed been scientifically successful, they should not have been wrong that many times with the true probability $P$- one. 
\bigskip

\textbf{Remark 2} We note that these two questions are different, especially because languages are creative. Note that any \textit{new} sentences can be created for the purpose of actual usage to effectively communicate with, even if they have never been circulated before. Thus, any of the potentially new sentences, which have not yet circulated in our world, should have zero frequency in the given ideal corpus, no matter how large we set the ideal corpus. Then, any of those languages should have a potentially infinite sequence of strings. Thus, the probability of obtaining any arbitrary new sentences simply becomes zero when we calculate its probability from the empirical distribution in the given ideal corpus. But it is obvious that the probability of generating any new sentence cannot be zero if we admit that languages are indeed creative. Thus, the probability of obtaining an arbitrary new language must be defined in a different way from the way of determining the probability of obtaining any given sentence in the ideal corpus. In any case, however, what we emphasize here is that this zero-frequency problem of creative language is not relevant to the case of a simple N-gram model where inputs are given. 

Recall that the unambiguous form of any inputs which have ever been given to machines in the tasks must be always included in the ideal corpus. For example, in speech recognition tasks, the unambiguous form of any given input must actually have been used for speech to effectively communicate before being given to machines, and therefore must be included in the ideal corpus at any given time $t$. Then, the problem of zero-frequency of newly created sentences in the ideal corpus does not apply here. Even in the extreme case where any input was just created at the very moment when the given speeches were made, the true probability of obtaining such sentences would not be definitely zero. 

In any case, any further detailed discussions on the creative language will be beyond the scope of this paper. Instead, we only emphasize here that what we try to show with Theorem 9 below is the following: If the true probability of obtaining any \textit{given} sentence can be conceptually defined by the empirical frequencies of the ideal corpus in Definition 5, the true probability in this simple N-gram model is an example of directly observable probability by Definition 7 and thus it is learnable.
\bigskip

\textbf{Remark 3} Due to the incredible technological development such as in the deep learning and big data algorithms, real machines may be able to obtain the empirical distributions of the entire population efficiently enough in the near future even if the size of the population may be astronomical. In these cases, the machines do not have to estimate the true distribution of the \textit{population} from some small samples. Instead, they only have to \textit{directly observe} the true probability from the empirical distribution of the entire population data.

For example, let us consider an imaginary case where the Wonka Factory announces a contest in which a Golden Ticket is included in each of the five random Wonka chocolate bars worldwide and the winners will receive the full package of prizes along with the lifetime supply of Wonka bars. Now, let us further assume that to gratify his daughter's desire to win one ticket, the father of Veruca Salt hires workers. Among them are included the data scientists who use machine learning algorithms to collect and then process all the necessary data about the entire number of the Wonka bars being circulated in the global market and the number of the tickets already claimed worldwide in order to calculate the true probability of winning a ticket at every instant. Now, it is clear in this case that machines used by the data scientists \textit{can }be thought to \textit{learn} the true objective probability of Veruca Salt's winning a Golden Ticket.

Just like in the Wonka Bars example, when machines \textit{directly observe} the empirical distribution of the \textit{true population} while using deep learning and big data algorithms for instance, ($i$) machines come to calculate it by the following \textit{definite and explicit} \textit{instructions} on how to \textit{effectively calculate} the true probability: collect all the necessary data from the \textit{given} whole population and then do the relevant calculation with the data, in order to obtain the relative frequencies of the target attributes in the entire population. ($ii$) Now, this effective calculation must be \textit{successful}, because the true probability \textit{is defined} to come from such an empirical distribution by Definition 7. Then, by Definition 3, machines come to \textit{learn} the true probability, because machines achieve computational success by ($i$) and ($ii$). Therefore, we argue that at least in such cases machines, in principle, can learn the true objective probability.
\bigskip

%%%%%%%%%%%%%%%%%%%%%%%%%%%%%%%%%%%%%%%%%%%%%%%%%%%%%%%%%%%%

\newpage
\section*{NeurIPS Paper Checklist}

\begin{enumerate}

\item {\bf Claims}
    \item[] Question: Do the main claims made in the abstract and introduction accurately reflect the paper's contributions and scope?
    \item[] Answer: \answerYes{} % Replace by \answerYes{}, \answerNo{}, or \answerNA{}.
    \item[] Justification: \ We did our best to reflect the paper's contributions and scope accurately in the abstract and introduction.

\item {\bf Limitations}
    \item[] Question: Does the paper discuss the limitations of the work performed by the authors?
    \item[] Answer: \answerYes{}
    \item[] Justification: \ We discuss the limitations of our work at several points in the paper.

\item {\bf Theory Assumptions and Proofs}
    \item[] Question: For each theoretical result, does the paper provide the full set of assumptions and a complete (and correct) proof?
    \item[] Answer: \answerYes{} % Replace by \answerYes{}, \answerNo{}, or \answerNA{}.
    \item[] Justification: \ We provide the full set of assumptions and a complete proof either in the main body of our paper or in the Appendix.

    \item {\bf Experimental Result Reproducibility}
    \item[] Question: Does the paper fully disclose all the information needed to reproduce the main experimental results of the paper to the extent that it affects the main claims and/or conclusions of the paper (regardless of whether the code and data are provided or not)?
    \item[] Answer: \answerNA{} % Replace by \answerYes{}, \answerNo{}, or \answerNA{}.
    \item[] Justification: \ Our paper does not include experiments.

\item {\bf Open access to data and code}
    \item[] Question: Does the paper provide open access to the data and code, with sufficient instructions to faithfully reproduce the main experimental results, as described in supplemental material?
    \item[] Answer: \answerNA{} % Replace by \answerYes{}, \answerNo{}, or \answerNA{}.
    \item[] Justification: \ Our paper does not include experiments requiring code.

\item {\bf Experimental Setting/Details}
    \item[] Question: Does the paper specify all the training and test details (e.g., data splits, hyperparameters, how they were chosen, type of optimizer, etc.) necessary to understand the results?
    \item[] Answer: \answerNA{} % Replace by \answerYes{}, \answerNo{}, or \answerNA{}.
    \item[] Justification: \ Our paper does not include experiments.

\item {\bf Experiment Statistical Significance}
    \item[] Question: Does the paper report error bars suitably and correctly defined or other appropriate information about the statistical significance of the experiments?
    \item[] Answer: \answerNA{} % Replace by \answerYes{}, \answerNo{}, or \answerNA{}.
    \item[] Justification: \ Our paper does not include experiments.

\item {\bf Experiments Compute Resources}
    \item[] Question: For each experiment, does the paper provide sufficient information on the computer resources (type of compute workers, memory, time of execution) needed to reproduce the experiments?
    \item[] Answer: \answerNA{} % Replace by \answerYes{}, \answerNo{}, or \answerNA{}.
    \item[] Justification: \ Our paper does not include experiments.
    
\item {\bf Code Of Ethics}
    \item[] Question: Does the research conducted in the paper conform, in every respect, with the NeurIPS Code of Ethics \url{https://neurips.cc/public/EthicsGuidelines}?
    \item[] Answer: \answerYes{} % Replace by \answerYes{}, \answerNo{}, or \answerNA{}.
    \item[] Justification: \ We have reviewed the NeurIPS Code of Ethics and tried our best to conform with it while doing the research conducted in this paper.

\item {\bf Broader Impacts}
    \item[] Question: Does the paper discuss both potential positive societal impacts and negative societal impacts of the work performed?
    \item[] Answer: \answerNA{} % Replace by \answerYes{}, \answerNo{}, or \answerNA{}.
    \item[] Justification: \ Our paper discusses very abstract and foundational questions regarding the nature of machine learning. It is unlikely that work on these questions will have societal impact, whether positive or negative. 
    
\item {\bf Safeguards}
    \item[] Question: Does the paper describe safeguards that have been put in place for responsible release of data or models that have a high risk for misuse (e.g., pretrained language models, image generators, or scraped datasets)?
    \item[] Answer: \answerNA{} % Replace by \answerYes{}, \answerNo{}, or \answerNA{}.
    \item[] Justification: \ Our paper does not contain datasets or models, and accordingly does not pose any such risks.

\item {\bf Licenses for existing assets}
    \item[] Question: Are the creators or original owners of assets (e.g., code, data, models), used in the paper, properly credited and are the license and terms of use explicitly mentioned and properly respected?
    \item[] Answer: \answerNA{} % Replace by \answerYes{}, \answerNo{}, or \answerNA{}.
    \item[] Justification: \ Our paper does not use existing assets.

\item {\bf New Assets}
    \item[] Question: Are new assets introduced in the paper well documented and is the documentation provided alongside the assets?
    \item[] Answer: \answerNA{} % Replace by \answerYes{}, \answerNo{}, or \answerNA{}.
    \item[] Justification: \ Our paper does not release new assets.

\item {\bf Crowdsourcing and Research with Human Subjects}
    \item[] Question: For crowdsourcing experiments and research with human subjects, does the paper include the full text of instructions given to participants and screenshots, if applicable, as well as details about compensation (if any)? 
    \item[] Answer: \answerNA{} % Replace by \answerYes{}, \answerNo{}, or \answerNA{}.
    \item[] Justification: \ Our paper does not involve crowdsourcing nor research with human subjects.

\item {\bf Institutional Review Board (IRB) Approvals or Equivalent for Research with Human Subjects}
    \item[] Question: Does the paper describe potential risks incurred by study participants, whether such risks were disclosed to the subjects, and whether Institutional Review Board (IRB) approvals (or an equivalent approval/review based on the requirements of your country or institution) were obtained?
    \item[] Answer: \answerNA{} % Replace by \answerYes{}, \answerNo{}, or \answerNA{}.
    \item[] Justification: \ Our paper does not involve crowdsourcing nor research with human subjects.

\end{enumerate}

\end{document}